\pdfoutput=1

\documentclass[11pt]{article}

\usepackage[]{EMNLP2023}

\usepackage{times}
\usepackage{latexsym}
\usepackage{graphicx}

\usepackage[T1]{fontenc}

\usepackage[utf8]{inputenc}

\usepackage{microtype}

\usepackage{inconsolata}

\usepackage{hyperref}

\title{Mavericks at BLP-2023 Task 1: Ensemble-based Approach Using Language Models for Violence Inciting Text Detection}

\author{Saurabh Page\thanks{~ Equal contribution}~~, 
Sudeep Mangalvedhekar$\footnotemark[1]$~, 
Kshitij Deshpande$\footnotemark[1]$~, \\
{\bf Tanmay Chavan}$\footnotemark[1]$~ \and
{\bf Sheetal Sonawane}$\footnotemark[1]$~ \\
Pune Institute of Computer Technology, Pune \\
\texttt{\{saurabhpage1,sudeepm117,kshitij.deshpande7,chavantanmay1402\}@gmail.com,}\\
\texttt{sssonawane@pict.edu}}

\begin{document}
\maketitle
\begin{abstract}
This paper presents our work for the Violence Inciting Text Detection shared task in the First Workshop on Bangla Language Processing. Social media has accelerated the propagation of hate and violence-inciting speech in society. It is essential to develop efficient mechanisms to detect and curb the propagation of such texts. The problem of detecting violence-inciting texts is further exacerbated in low-resource settings due to sparse research and less data. The data provided in the shared task consists of texts in the Bangla language, where each example is classified into one of the three categories defined based on the types of violence-inciting texts. We try and evaluate several BERT-based models, and then use an ensemble of the models as our final submission. Our submission is ranked 10th in the final leaderboard of the shared task with a macro F1 score of 0.737. 
\end{abstract}

\section{Introduction}
\par In today's digital age, numerous social platforms play an important role in connecting individuals around the world. However, certain malicious elements resort to using these platforms to instigate riots, protests, and disturbances that lead to violence. The online posts and comments involve direct threats pertaining to resocialization, vandalism, and deportation while indirect threats involve derogatory language and abusive remarks. These texts which are thought to be a potential reason for instigating violence are called violence-inciting texts. Classifying them has become a major challenge and various techniques are used to implement it. These applications can be used to monitor social media websites and take precautions to avoid any mishaps. Thus the task boils down to text classification wherein we need to label such texts into predefined categories.
\par The shared tasks involve performing sentiment analysis and text classification. The BLP Workshop offers two shared tasks namely, Violence Inciting Text Detection (VITD) \cite{blp2023-overview-task1} and Sentiment Analysis of Bangla Social Media Posts \cite{blp2023-overview-task2}. Our team, under the name \textit{Mavericks} contested in the VITD task under the Codalab username \textit{kshitij}. Our paper illustrates work on the VITD task where we have to classify text into predefined categories of violence. The dataset consists of text in the Bangla language with a length of up to 600 words. 
\par Transformer-based models (\citet{vaswani2023attention}) such as BERT (\citet{devlin-etal-2019-bert}) have brought revolution in NLP-related tasks and have proved their worth by attaining state-of-the-art (SOTA) results on several benchmarks \cite{lan2020albert}. Large Language Models (LLMs) are increasingly used for text classification tasks \cite{liu2019roberta}. We use several transformer-based pre-trained models to achieve higher performance. Furthermore, we use ensembling techniques to produce better results. We present our results after experimenting with several models and ensembling techniques.

\section{Related Work}
\citet{pang2002thumbs} considers classifying documents by overall sentiment and not just by topic. The three machine learning methods - Naive Bayes, Maximum Entropy Classification, and Support Vector Machines did not perform well on sentiment analysis. \citet{warner-hirschberg-2012-detecting} describes the definition of hate speech as the collection and annotation of hate speech corpus along with a mechanism for detecting some commonly used methods of evading common “dirty word” filters. \citet{6975602} automatically detects threats related to violence using machine learning methods. 24,840 sentences obtained from YouTube comments were manually annotated and were used to train and test the machine learning model. They suggest that the features that combine main words and the distance between those in the sentence attain the best results. 

\par \citet{hassan2016sentiment} provides a textual dataset in Bangla and Romanized Bangla language which can be directly used for sentiment analysis. The dataset was tested using Long Short Term Memory (LSTM) a type of Deep Recurrent Model. Two types of loss functions are used- binary cross-entropy and categorical cross-entropy. \citet{8843606} used Linear Support Vector Classifier (LinearSVC), Logistic Regression (Logit), Random Forest (RF), Artificial Neural Network (ANN), and Recurrent Neural Network with an LSTM cell. A Deep-learning-based algorithm using RNN beats all other algorithms by gaining the highest accuracy 82.20\%. 

\par In 2017, “Attention is all you need”(\citet{vaswani2023attention}) introduces the concept of Transformers which transformed the Natural Language Processing (NLP) landscape. The paper introduced the concept of self attention. In 2019, a new language model called BERT (Bidirectional Encoder Representations from Transformers) was put forward by \citet{devlin-etal-2019-bert}. BERT is designed to pre-train deep bidirectional representations from the unlabelled text by joint conditioning on both the left and right context in all layers. Pre-trained BERT can be used for numerous tasks including text classification by fine-tuning it.

\par \citet{10.1145/3575882.3575935} proposes a BERT-based method for Aspect-based Sentiment Analysis that can identify and handle conflicting opinions. The method achieves better results on three-class and four-class classification tasks. \citet{10103337} performs sentiment analysis of book reviews in Bangla. A dataset consisting of 5189 reviews was produced by crawling data. An investigation of several deep neural network models and three transformer models is performed. XLM-R outperforms all models, achieving a weighted F1-score of 88.95\% on the test data. \citet{anan2023interpretable} performs sarcasm detection using BERT and achieved 99.60\% accuracy. A new dataset "BanglaSarc", consisting of comments from Facebook and YouTube was used. \citet{s22114157} utilizes a deep integrated model "CNN-BiLSTM" for enhanced performance of decision-making in text classification. 

\begin{figure*}[t]
    \centering
    \includegraphics[width=\textwidth]{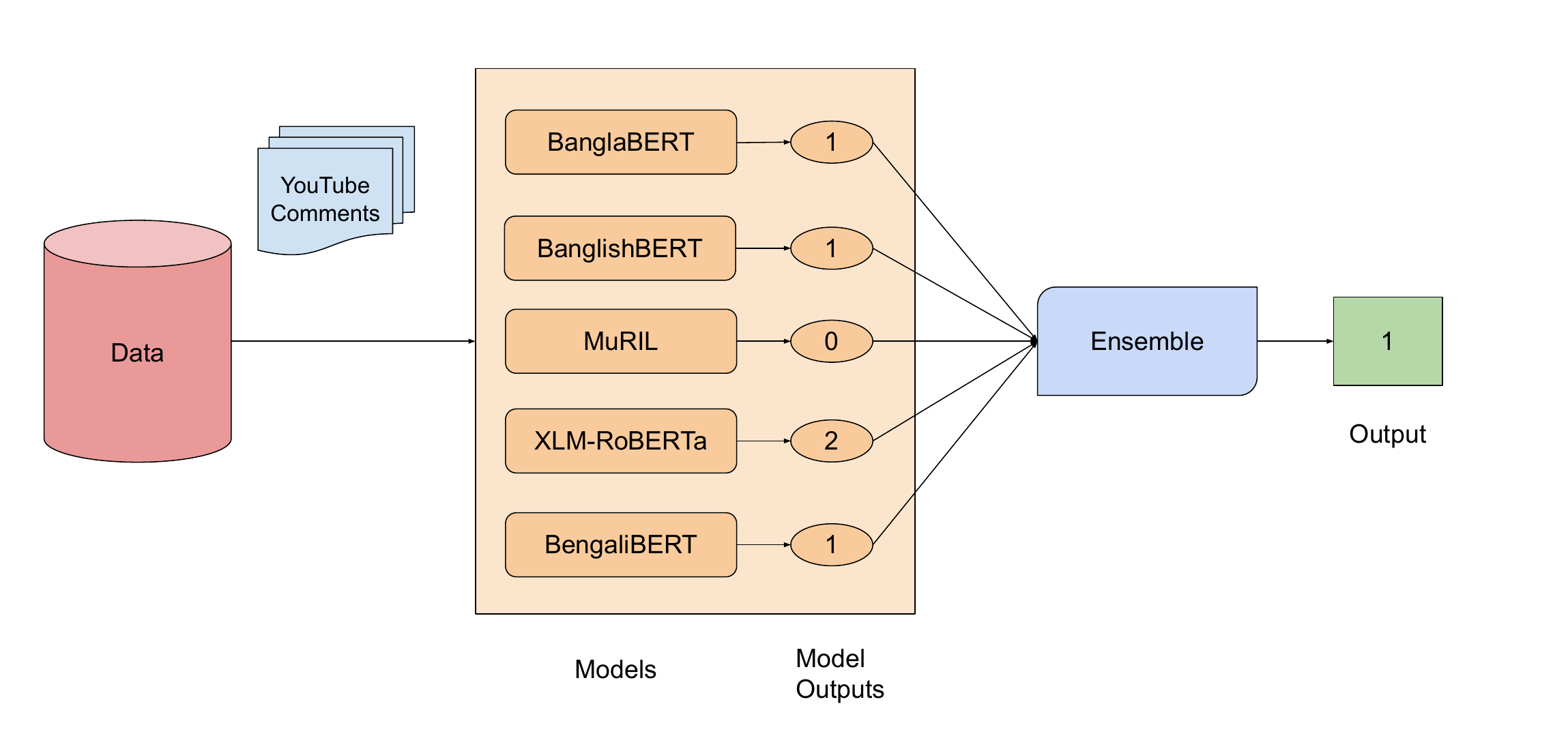}
    \caption{System Architecture}
    \label{fig:ensemble}
\end{figure*}

\section{Data}
We use the \textit{Vio-Lens} dataset provided by \citet{SahaAndJunaed} for the task. The dataset consists of YouTube comments related to nine violent incidents in the Bengal region (Bangladesh and West Bengal) within the past ten years. The comments are in the Bangla language with a length of up to 600 words. The dataset consists of two attributes: text, and label. The "text" column contains comments while the "label" column contains three values 0, 1, and 2 representing Non-Violence, Passive Violence, and Direct Violence respectively. The training dataset consists of 2700 samples out of which approximately 15\% depict direct violence, 34\% portray passive violence and the remaining 51\% represent non-violent instances. The development dataset consists of 1330 samples out of which approximately 15\% illustrate direct violence, 31\% depict passive violence and the remaining 54\% represent non-violent instances. The test dataset provided at the time of evaluation consists of 2016 samples as seen in Table \ref{Table:1}.
\begin{table}[ht]
\begin{center}
\begin{tabular}{|c|c|}
\hline
\textbf{Dataset} & \textbf{Number of Samples} \\ \hline
Training & 2700 \\ \hline
Development & 1300 \\ \hline
Testing & 2016 \\ \hline
\end{tabular}

\caption{Dataset statistics.}
\label{Table:1}
\end{center}
\end{table}

\section{System}
This shared task discusses the problem of Violence Inciting Text Detection. This issue falls under the category of classification, for which transformer-based models have seen extensive application and have demonstrated outstanding performance. As a result, we use and experiment with a variety of such models and ensembling techniques in our research. In the section below, the approaches have been briefly discussed.

\subsection{BERT-based Models}
\citet{khanuja2021muril} discusses how even the state-of-the-art models do not perform satisfactorily well in Indian languages and summarises the gaps found. To mitigate these gaps, they propose their model "MuRIL"\footnote{Model link: \href{https://huggingface.co/google/muril-base-cased}{https://huggingface.co/google/muril-base-cased}} which is trained in 16 different Indian languages and English. As we deal with the Bangla language in this task, MuRIL is specifically relevant. It is trained on two learning objectives, first - Masked  Language Modeling, and second - Translation Language Modeling. The model has 236M parameters and a vocabulary of 197285.

\citet{joshi2022l3cubehind} states that even though multilingual BERT models are suitable for very low-resource languages, models trained on a single language outperform it when sufficient resources for a language are available. Based on this assertion, they propose several models for different languages. Bengali-BERT\footnote{Model link: \href{https://huggingface.co/l3cube-pune/bengali-bert}{https://huggingface.co/l3cube-pune/bengali-bert}} is of specific interest to us. Existing multi-lingual models are fine-tuned on the Bangla language corpus to create this model.

\citet{DBLP:journals/corr/abs-1911-02116} demonstrates how cross-lingual understanding can be improved by pre-training multilingual models on a large scale. XLM-RoBERTa\footnote{Model link: \href{https://huggingface.co/xlm-roberta-base}{https://huggingface.co/xlm-roberta-base}} is pre-trained on 2.5TB of filtered CommonCrawl data, which included 100 different languages. It is trained with the multilingual Masked Language Modeling objective. We use the base-sized model in our experiments, XLM-RoBERTa-base which has 270M parameters.

\subsection{ELECTRA-based Models}
In \citet{bhattacharjee-etal-2022-banglabert}, authors propose training Large Language Models on a dataset specifically tailored for pre-training transformer models useful for Natural Language Processing tasks in the Bangla language. The authors observe that instead of the Masked Language Modeling(MLM) pre-training approach used to train BERT-based models, using ELECTRA and its Replaced Token Detection (RTD) objective provides significant performance improvements while at the same time using significantly less compute power for pre-training. Two Large Language Models are pre-trained namely BanglaBERT\footnote{Model link: \href{https://huggingface.co/csebuetnlp/banglabert}{https://huggingface.co/csebuetnlp/banglabert}} and BanglishBERT\footnote{Model link:\\ \href{https://huggingface.co/csebuetnlp/banglishbert}{https://huggingface.co/csebuetnlp/banglishbert}}.
\par The dataset used for pre-training these models was collected by crawling 110 Bangla websites. The total size of the dataset is 27.5GB consisting of 5.25 million documents.
\par BanglaBERT, introduced in \citet{bhattacharjee-etal-2022-banglabert}, is trained using the ELECTRA pre-training approach consisting of a 12 layer Transformer encoder with 768 embedding size and 12 attention heads. The batch size used is 256 and it is trained for a total of 2.5M steps.
\par BanglishBERT introduced in \citet{bhattacharjee-etal-2022-banglabert}, is a bilingual model trained on Bangla and English data. It acts as the generator model in the pre-training phase of the ELECTRA approach. BERT pretraining corpus is used along with Bangla data which is upsampled to have equal participation of both languages.
\par BanglaBERT outperforms other multilingual models such as mBERT and XLM-R (base) on a Bangla-specific benchmark introduced by the authors - Bangla Language Understanding Benchmark (BLUB). BanglaBERT achieves impressive results while having better convergence and thus being more compute-efficient than other previously pre-trained multilingual models.

The batch size used for training all the models is 16. The learning rate used is 1\textit{e}-5. We use the AdamW optimizer and the Cross-Entropy Loss. We train the models for 10 epochs. All of the models we use in the experiments are freely available on HuggingFace. We have tagged the models with their respective HuggingFace model links in the footnotes. We use the tokenizers recommended by the model developers provided along with the HuggingFace models.

\begin{table}[ht]
\begin{center}
\begin{tabular}{|c|c|c|}
\hline
\textbf{Model} & \textbf{\begin{tabular}[c]{@{}c@{}}Pre-Training \\ Approach\end{tabular}} & \textbf{\begin{tabular}[c]{@{}c@{}}Macro \\ F1-\\ Score\end{tabular}} \\ \hline
\textbf{BanglaBERT} & \textbf{\begin{tabular}[c]{@{}c@{}}ELECTRA \\ (RTD)\end{tabular}} & \textbf{0.791} \\
BanglishBERT & \begin{tabular}[c]{@{}c@{}}ELECTRA \\ (RTD)\end{tabular} & 0.742 \\
MuRIL & MLM & 0.753 \\
XLM-RoBERTa & MLM & 0.743 \\
BengaliBERT & MLM & 0.739 \\ \hline
\textbf{\begin{tabular}[c]{@{}c@{}}Ensemble - Hard\\ Voting\end{tabular}} & \textbf{-} & \textbf{0.782} \\ \hline
\end{tabular}
\caption{Results on the development dataset.}
\label{Table:2}
\end{center}
\end{table}

\section{Ensembling}
Ensembling is a technique that combines the results of various models to generate the system's eventual result. Statistical as well as non-statistical methods are used for this purpose. Ensembling is useful as it helps generate results that are better than the results given by the individual models. \\
Amongst several methods leveraged for ensembling, we use the "hard voting" ensemble technique. In hard voting, the majority vote or the "mode" of all the predictions is selected as the final prediction. It helps improve the robustness of the system and minimizes the variance in the results. The ensembling mechanism is illustrated in figure \ref{fig:ensemble}. \\\\
In the post-evaluation phase, we experiment with the weighted ensemble keeping in mind the varied performances of the underlying models. We give higher weights to the models which perform better. We experiment with different weights for models and choose the weights which provide the best results. We also explore different subsets of the 5 mentioned models and form an ensemble of the models to generate predictions. However, the ensembles of the subsets did not provide improvements to our system's predictions.

\begin{table}[ht]
\begin{center}
\begin{tabular}{|c|c|c|}
\hline
\textbf{Model} & \textbf{\begin{tabular}[c]{@{}c@{}}Pre-Training \\ Approach\end{tabular}} & \textbf{\begin{tabular}[c]{@{}c@{}}Macro \\ F1-\\ Score\end{tabular}} \\ \hline
\textbf{BanglaBERT} & \textbf{\begin{tabular}[c]{@{}c@{}}ELECTRA \\ (RTD)\end{tabular}} & \textbf{0.733} \\
BanglishBERT & \begin{tabular}[c]{@{}c@{}}ELECTRA \\ (RTD)\end{tabular} & 0.662 \\
MuRIL & MLM & 0.720 \\
XLM-RoBERTa & MLM & 0.705 \\
BengaliBERT & MLM & 0.690 \\ \hline
\textbf{\begin{tabular}[c]{@{}c@{}}Ensemble - Hard\\ Voting\end{tabular}} & \textbf{-} & \textbf{0.737} \\ \hline
\textbf{\begin{tabular}[c]{@{}c@{}}Weighted\\ Ensemble\end{tabular}} & \textbf{-} & \textbf{0.745} \\ \hline
\end{tabular}
    
\caption{Results on the test dataset.}
\label{Table:3}
    
\end{center}
\end{table}

\section{Results}
\par This section discusses the findings of our experiments. Table \ref{Table:3} contains our results for the models and ensembles. The macro F1 score is the shared task's official score statistic for the Violence Inciting Text Detection task.
\par BanglaBERT achieves the best result with a macro F1 score of 0.733 among the individual models as seen in table \ref{Table:3}. This performance can be attributed to the fact that BanglaBERT is trained on a carefully curated dataset of the Bangla language, unlike other multi-lingual models such as MuRIL and XLM-RoBERTa whose training corpus consists of numerous other languages. It also uses the ELECTRA approach for pre-training which involves using the Replaced Token Detection (RTD) objective instead of the Masked Language Modeling (MLM) objective used in other multilingual BERT models; this allows BanglaBERT to achieve a better performance whilst also converging faster. The performance of MuRIL and XLM-RoBERTa is limited by the quantity and quality of Bangla text they used in pre-training, although it is worth noting that the models will perform much better in a multilingual setting.
\par BanglaBERT performs marginally better on the development dataset as seen in Table \ref{Table:2}, than the ensemble of the five mentioned models but underperforms on the test dataset. We can attribute this slight difference to variations in the performance of individual models on different data samples and the ensemble's stable and high performance across different data samples. We chose ensembling as the final approach for our final submission owing to its better generalizability and low variance in its predictions. Our final submission to the task using the hard voting ensembling mechanism achieves a macro F1 score of 0.737. 
\par Our post-evaluation phase experiments yielded better results with the weighted ensembling technique. The weighted ensemble achieves a macro F1 score of 0.745 on the test dataset, thus outperforming the hard voting-based ensembling approach. 

\section{Conclusion}
We present our approach for the shared task in the First Workshop on Bangla Language Processing through this paper. We experiment with several BERT and ELECTRA-based models as a part of our efforts. We observed that the ELECTRA-based BanglaBERT model has the best performance, followed by MuRIL. We can see that the ELECTRA-bSased models have similar performances compared to their BERT-based counterparts, despite being smaller in size. Our final submission consists of predictions generated by ensembling the evaluated models and has a macro F1 score of 0.737, placing us tenth on the shared task leaderboard. Our experiments have shed light on several further avenues for improvement. Larger pre-training datasets are required for better low-resource models. More sophisticated ensembling techniques can better utilize the performance of individual models and need to be researched further.

\section*{Acknowledgment}

We thank the Pune Institute of Computer Technology's  Computational Linguistics Research Lab for providing us guidance and support necessary for the work. We are grateful for their help.

\section*{Limitations}
The models that have been utilized are compute-intensive and thus can pose a challenge in real-world applications. Also, it must be considered that the pre-training and evaluation datasets, although of high quality, might possess certain implicit biases and thus might not fully model real-world situations.

\bibliography{anthology,custom}
\bibliographystyle{acl_natbib}

\end{document}